\documentclass{article}

\usepackage{graphicx}
\usepackage{booktabs} 
\usepackage{mathtools}
\usepackage{comment}
\usepackage{soul}
\usepackage{courier}
\usepackage[numbers]{natbib}
\usepackage{todonotes}
\usepackage{algorithm}
\usepackage{multirow}
\usepackage{authblk}
\usepackage{amsmath,amssymb,amsfonts}
\usepackage{algorithmic}
\usepackage{textcomp}
\usepackage{xcolor}
\usepackage{xspace}
\usepackage{soul,color}  
\usepackage{ragged2e}
\usepackage{makecell}
\usepackage[figurename=Fig.]{caption}
\usepackage{tikz}
\usepackage{fancyhdr}
\usepackage[normalem]{ulem}
\usepackage{flushend}
\usepackage[utf8]{inputenc} 
\usepackage[T1]{fontenc}    
\usepackage{url}            
\usepackage{booktabs}       
\usepackage{amsfonts}       
\usepackage{nicefrac}       
\usepackage{microtype}      
\usepackage{xcolor}         
\usepackage{changepage,threeparttable} 
\usepackage{letltxmacro}
\newcommand{\niparagraph}[1]{\vspace{1pt}\noindent\textbf{#1}}
\usepackage{enumitem}

\usepackage[bookmarks=true,breaklinks=true,letterpaper=true,colorlinks,linkcolor=blue,citecolor=blue,urlcolor=blue]{hyperref}

\newcommand{\PCignore}[1]{}
\def\Snospace~{\S{}}

\renewcommand\figurename{Fig.}

\newcommand{\squishlist}{
 \begin{list}{$\bullet$}
  { \setlength{\itemsep}{0pt}
     \setlength{\parsep}{3pt}
     \setlength{\topsep}{3pt}
     \setlength{\partopsep}{0pt}
     \setlength{\leftmargin}{1.5em}
     \setlength{\labelwidth}{1em}
     \setlength{\labelsep}{0.5em} } }

\newcommand{\squishlisttwo}{
 \begin{list}{$\bullet$}
  { \setlength{\itemsep}{0pt}
     \setlength{\parsep}{0pt}
    \setlength{\topsep}{0pt}
    \setlength{\partopsep}{0pt}
    \setlength{\leftmargin}{2em}
    \setlength{\labelwidth}{1.5em}
    \setlength{\labelsep}{0.5em} } }

\newcommand{\squishend}{
  \end{list}  }

\newcommand{\alg}{{Mask-Decay}\xspace}

\usepackage[final]{neurips_2022}

\title{Training Recipe for N:M Structured Sparsity\\with Decaying Pruning Mask}

\newcommand{\MYhref}[3][blue]{\href{#2}{\color{#1}{#3}}}%

\author{
\textbf{Sheng-Chun Kao}$^*$, \textbf{Amir Yazdanbakhsh}$^{*\dagger}$, \textbf{Suvinay Subramanian}$^{\ddagger}$\\\textbf{Shivani Agrawal}$^{\dagger}$, \textbf{Utku Evci}$^{\dagger}$, \textbf{Tushar Krishna}\\
Georgia Tech\hspace{0.5em}$^{\dagger}$Google Research, Brain Team\hspace{0.5em}$^{\ddagger}$Google\\
\texttt{\footnotesize{\MYhref{mailto:felix@gatech.edu}{felix@gatech.edu}, \MYhref{mailto:ayazdan@google.com}{ayazdan@google.com}, \MYhref{mailto:suvinay@google.com}{suvinay@google.com}}}\\\texttt{\footnotesize{\MYhref{mailto:shivaniagrawal@google.com}{shivaniagrawal@google.com}, \MYhref{mailto:evcu@google.com}{evcu@google.com}, \MYhref{mailto:tushar@ece.gatech.edu}{tushar@ece.gatech.edu}}}\\
(\footnotesize{$^*$Equal Contribution})
}

\begin{document}
\maketitle
\begin{abstract}
Sparsity has become one of the promising methods to compress and accelerate Deep Neural Networks (DNNs).
Among different categories of sparsity, structured sparsity has gained more attention due to its efficient execution on modern accelerators.
Particularly, N:M sparsity is attractive because there are already hardware accelerator architectures that can leverage certain forms of N:M structured sparsity to yield higher compute-efficiency.
In this work, we focus on N:M sparsity and extensively study and evaluate various training recipes for N:M sparsity in terms of the trade-off between model accuracy and compute cost (FLOPs).
Building upon this study, we propose two new decay-based pruning methods, namely ``pruning mask decay'' and ``sparse structure decay''.
Our evaluations indicate that these proposed methods consistently deliver state-of-the-art (SOTA) model accuracy, comparable to unstructured sparsity, on a Transformer-based model for a translation task.
The increase in the accuracy of the sparse model using the new training recipes comes at the cost of marginal increase in the total training compute (FLOPs).
\end{abstract}
\section{Introduction}

Deep Neural Networks (DNNs) have shown success in many domains such as computer vision, language modeling, machine translation, and so on. An trend of SOTA DNN models is that the model size increases quickly with time. For example T5 from Google~\cite{t5_model}, OPT from Meta~\cite{meta_opt} and GPT-3 from OpenAI~\cite{gpt3} have over 100 billions parameters, making them hard to be deployed and inaccessible for many practitioners with limited compute resources. Another line of effort in the DNN community is to propose different methods to compress the models, such as quantization~\cite{shen2020q,kim2021bert,zafrir2019q8bert,zhang2020ternarybert,yazdanbakhsh2018releq}, sparsification~\cite{evci2019difficulty,han2015deep, guo2016dynamic,he2017channel,molchanov2016pruning,yao2019balanced,zhu2017prune,gamboa2020campfire,narang2017exploring,narang2017block,elsen2020fast,park2018squantizer,kalchbrenner2018efficient,evci2020rigging}, and distillation~\cite{sanh2019distilbert,jiao2019tinybert,sun2020mobilebert,wang2020minilm}.

In this paper, we focus on sparsification (or pruning), which prunes a portion of the parameters in the model by setting their values to 0. It can reduce the amount of compute by skipping multiplications with 0, reduce  memory usage by using compressed sparse representations such as COO, CSR, and so on~\cite{qin2021extending}, and save energy/power by reducing memory accesses and computations. It opens up the possibility of deploying a large model in resource-limited devices. However, sparsification is often about trading-off between model quality\footnote{In this paper, we refer to algorithmic-wise criteria such as accuracy, recall, and precision as \textit{model quality}; we refer to model runtime/latency as \textit{model performance}.} and compression ratio. For example, many studies show promising results in sparsifying image classification models to around 90\%-95\% sparsity (5\%-10\% density) without quality loss~\cite{guo2016dynamic, han2015learning}. With the success of Transformers in natural language processing, there is rising interest in investigating sparsification in Transformer models, where around 80\%-90\% sparsity can be achieved. Sparsification in language models has huge potential benefits especially in encoder-decoder tasks such as translation. Since decoder needs to be run iteratively for all N tokens in a sequence, even minor performance improvements in the decoder can improve performance significantly. In this paper, we demonstrate our method with an encoder-decoder Transformer-based translation model.

While sparsification can effectively reduce the memory requirement, generally leveraging induced (unstructured) sparsity in the model for higher performance improvements is challenging.
The irregularity of the sparsity pattern makes it challenging to be effectively leveraged by the dense accelerators such as GPU and TPU.
The sparsified models often ends up with similar or worse performance (because of the extra complexity to compress and decompress the parameters) than their dense counterparts~\cite{gpu_ampere,non_structured_pruning,renda2020comparing,he2017channel,lin2021filter,gamboa2020campfire,zhu2019sparse, wen2016learning,evci2020gradient}. 

To this end, structured sparsity, which regularizes the sparsity pattern such as channel/filter sparsity~\cite{li2016pruning,wen2016learning,he2017channel}, or block sparsity~\cite{non_structured_pruning, tan2020pcnn}, have become increasingly popular owing to their hardware-friendly nature.
For example the dense accelerator can skip a full channel computation when it is sparsified without any low-level modification. The caveat is structured sparsity also introduces larger quality loss. Recent research~\cite{yao2019balanced, kang2019accelerator} found fine-grained N:M structured sparsity, which keeps N out of consecutive M elements in the the weight tensor, can ameliorate the quality loss. Moreover, with the launch of 2:4 structured-sparse tensor core in GPU Ampere architecture~\cite{gpu_ampere} developing sparse training recipes for N:M sparsity has acquired increased interest~\cite{pool2021channel,mishra2021accelerating,nvidia_asp, sr_ste}.

In this paper, we demonstrate a training recipe for N:M structure sparsity in Transformer-based translation task and propose two techniques. We propose \textit{Structure Decay} an iterative pruning approach tailored for N:M sparsity. We propose \textit{Mask Decay}, which gradually decays the mask
from 1, to 0.9, 0.8, .., to 0, instead of the conventional 1/0 mask. We found these techniques can stabilize the training and achieve better quality and compression rate. We make following contributions:

\squishlist
\item We compare \textit{Structure Decay} and \textit{Mask Decay} with the state-of-the-art N:M sparsity training recipes, SR-STE~\cite{sr_ste}. They achieve (geomean) 0.004 and (geomean) 0.006 accuracy improvement over SR-STE, respectively.
\item \textit{Mask Decay} enables ``structured pruning'' to achieve comparable quality and compression rate to ``unstructured pruning''.
\squishend
\begin{figure*}
\begin{center}
\includegraphics[width=1\linewidth]{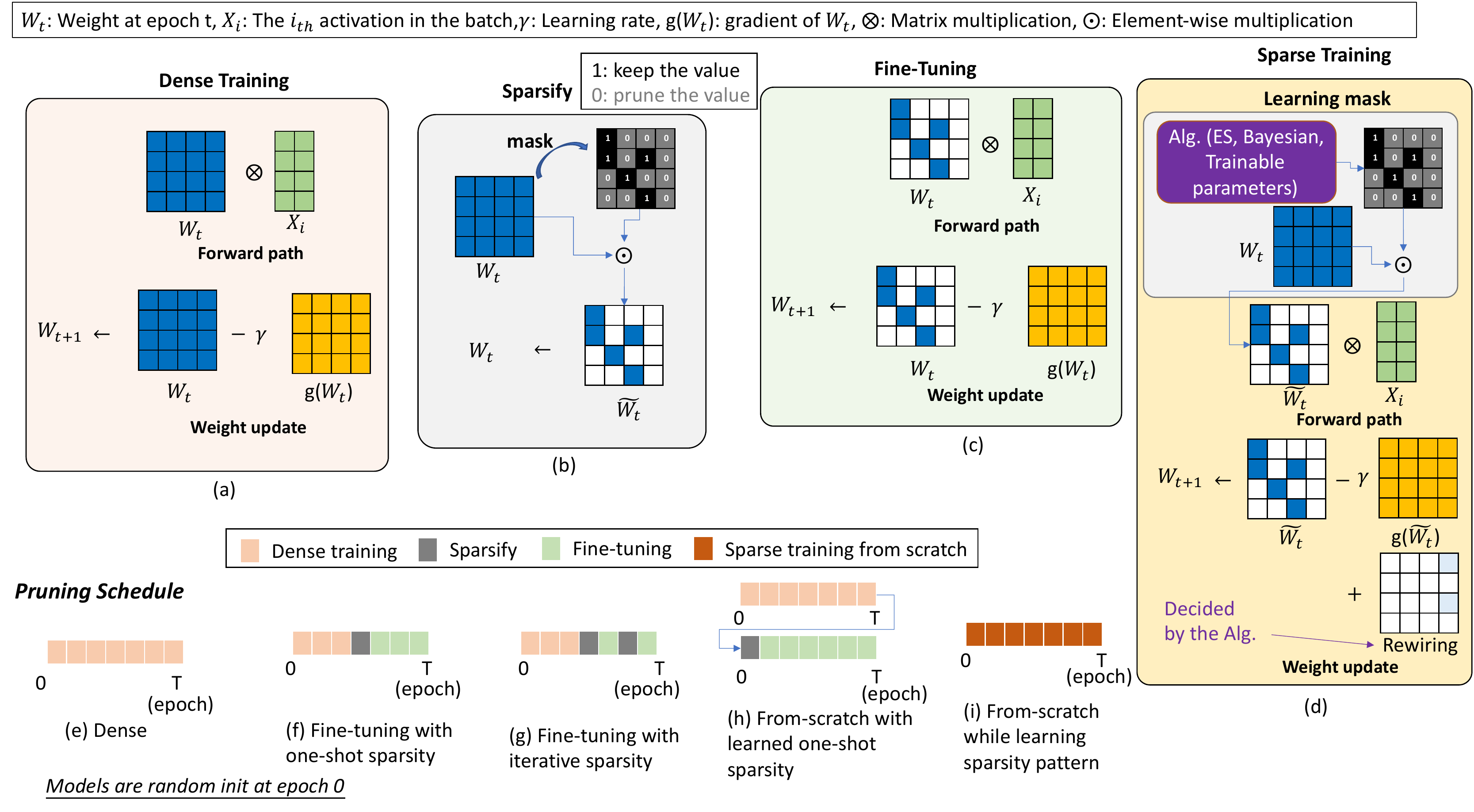}
\end{center}

\caption{The compute flows of (a) Dense training, (b) Sparsify, (c) Fine-tuning, and (d) Sparse training. The training schedule of (e) regular dense training, (f) fine-tuning with one-shot sparsifying, (g) fine-tuning with iterative sparsifying, (h) from-scratch with learned one-shot sparsity pattern, and (i) from-scratch while learning sparsity pattern. The sparsify algorithm in (d): ES such as \cite{mocanu2018scalable}, Bayesian Optimization such as \cite{bellec2017deep}, Trainable parameters such as \cite{wortsman2019discovering,dettmers2019sparse, kusupati2020soft}.}

\label{fig:dense_sparse}
\end{figure*}

\begin{figure*}
\begin{center}
\includegraphics[width=1\linewidth]{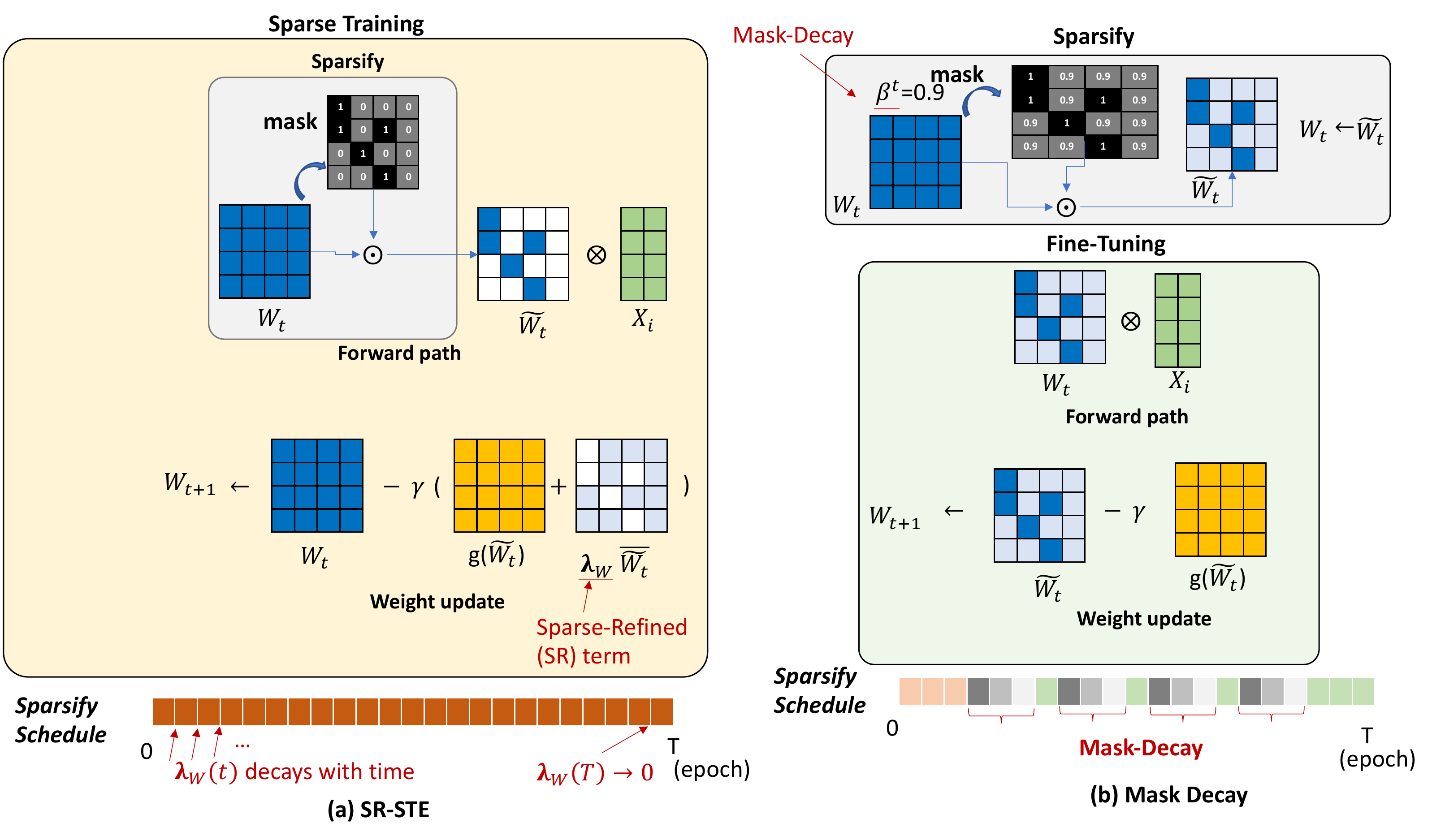}
\end{center}

\caption{The weight update scheme of (a) SR-STE~\cite{sr_ste} and (b) \alg.}

\label{fig:alg_flow}
\end{figure*}

\section{Related Work}
We primarily focus on weight sparsification in this work. A sparsification recipe includes: 1) pruning criteria, 2) pruning schedule, and 3) sparsity pattern.

\niparagraph{Pruning criteria.}
Pruning criteria is the criteria to decide which elements to prunes in the weight tensor. Magnitude pruning, which selects the pruning elements by their absolute values, is the most widely used method~\cite{renda2020comparing,guo2016dynamic,lee2018snip,frankle2018lottery,gale2019state,zhu2017prune,han2015deep,liu2018rethinking,pool2021channel,mishra2021accelerating}.
Some other metrics such as gradient-based~\cite{yeom2021pruning, evci2020rigging}, Hessian based~\cite{lecun1989optimal}, connection sensitivity~\cite{lee2018snip}, and salient-based~\cite{molchanov2019importance,lee2018snip} are also used. In this paper, we use magnitude pruning.

\niparagraph{Pruning schedule.}
There are coarsely four different pruning schedules. \textit{1) Fine-tuning with one-shot pruning} (\autoref{fig:dense_sparse}f)~\cite{mishra2021accelerating,pool2021channel, frankle2018lottery,lee2018snip}, which trains a dense models, prunes the weight with one-shot, and re-trains the model in order to recover the quality loss. \textit{2) Fine-tuning with iterative pruning} (\autoref{fig:dense_sparse}g)~\cite{evci2019difficulty,han2015deep, guo2016dynamic,he2017channel,molchanov2016pruning,yao2019balanced,zhu2017prune,gamboa2020campfire,narang2017exploring,narang2017block,elsen2020fast,park2018squantizer,kalchbrenner2018efficient,evci2020rigging}, which trains a dense model and then iterates between pruning and re-training. This schemes are usually found to has higher ability to recover the quality loss. \textit{3) From-scratch with learned one-shot pruning pattern} (\autoref{fig:dense_sparse}h)~\cite{frankle2020pruning,evci2019difficulty}, which determines the sparsity pattern from the trained dense version and trains a sparse model from scratch. \textit{4) From-scratch while learning sparsity pattern} (\autoref{fig:dense_sparse}i)~\cite{wortsman2019discovering,dettmers2019sparse,gale2019state,kusupati2020soft,evci2020rigging,bellec2017deep,mocanu2018scalable,sr_ste,evci2020gradient}, which trains a sparse model from scratch while learning sparsity patterns simultaneously.

\niparagraph{Sparsity pattern.} \textit{Unstructured Sparsity} prunes the model without any sparsity pattern constraint~\cite{renda2020comparing,guo2016dynamic,lee2018snip,frankle2018lottery,gale2019state,zhu2017prune,han2015deep,liu2018rethinking,wortsman2019discovering,dettmers2019sparse,gale2019state,kusupati2020soft,evci2020rigging,bellec2017deep,mocanu2018scalable}. It is often found to be able to prune the model size to an order of magnitude smaller while keeping the model quality. However, it has the challenge of similar or worse (because of the additional complexity) runtime than the dense model owing to its irregular sparsity pattern. \textit{Coarse-grained Structured Sparsity} constrains the pruning scheme to prune the model in a coarse-grained manner such as filter/channel pruning~\cite{li2016pruning,wen2016learning,he2017channel}, block-wise pruning~\cite{wen2016learning,non_structured_pruning,narang2017block,gray2017gpu}, and so on. By skipping the full computation at a coarse-granularity of the computation, this scheme can often achieve speedup in dense computation accelerators such as GPUs and TPUs; however this often sacrifices some model quality. These studies often trade off between performance and quality for different application needs. \textit{Fine-grained N:M Structure Sparsity}, which prunes (M-N) out of consecutive M elements. Some early works rely on special threading and grouping techniques~\cite{yao2019balanced} or specialized sparse accelerators~\cite{kang2019accelerator} to leverage this fine-grained pattern. With the 2:4 structured-sparse GEMM support in tensor cores in GPU Ampere architecture~\cite{gpu_ampere}, many recent works start to investigate in different training recipe for N:M sparsity pattern to leverage the existing hardware~\cite{pool2021channel,mishra2021accelerating,nvidia_asp,sr_ste}.

\niparagraph{Proposed recipe and the SOTA SR-STE~\cite{sr_ste}}.
In this paper, our sparse training recipe is (Pruning criteria: magnitude pruning, Pruning schedule: fine-tuning with iterative pruning, Sparsity pattern: fine-grained N:M structure sparsity). The recipe for SR-STE~\cite{sr_ste} is (Pruning criteria: magnitude pruning, Pruning schedule: from-scratch with learning iterative pruning, Sparsity pattern: fine-grained N:M structure sparsity). Our methods, Mask Decay and Structure Decay, are techniques to improve the training quality of ``fine-tuning with iterative pruning''. SR-STE~\cite{sr_ste} is a ``Sparse-Refined (SR)'' technique to stabilize the training of ``from-scratch with learning iterative pruning''. Both SR-STE~\cite{sr_ste} and us are proposing techniques to pursue high quality sparsification for fine-grained N:M structure sparsity pattern.
\section{Methodology}
\begin{table}[ht]\centering
\caption{The compute and memory contributions of the three major layers in Transformers. Einsum: computation of attention scores and weighted sum of values by the attention scores. Projections: projecting inputs to key, query, and value, and projection weighted sum of values to outputs. Feed Forward: The multiple feed forward layers at the end of the attention layer. The other layers/ operations such as ReLU, LayerNorm, Add, Softmax, embedding, and so on have little contributions to the FLOPS and parameters of the Transformers, hence not included in this estimation. The feed forward layers account for around 64\% of overall FLOPs and 67\% of parameters. These estimations are made under model configuration in \autoref{table:setup}.}
\scriptsize
\begin{tabular}{lrrrrrrr}\toprule
&Einsum &Projections &Feed Forward &Einsum &Projections &Feed Forward \\\midrule
(T)FLOPS &1.6 &13.2 &26.4 &4\% &32\% &64\% \\
Params (MB) &0.0 &50.3 &100.7 &0\% &33\% &67\% \\
\bottomrule
\end{tabular}
\label{table:compute_contribution}
\vspace{0.4cm}
\end{table}

\niparagraph{Workload and model.}
We evaluate different sparsification methods on the WMT translate task~\cite{wmt_dataset} that uses a Transformer-based model~\cite{vaswani2017attention}, and is a key benchmark in machine translation research. They hold several translation datasets across different languages. 
The encoder and decoder blocks in this model each have six attention layers with 16 heads. The embedding dimension for both input and query/key/value are 1024. The feed-forward blocks within each attention head has 4096 neurons.
For all the experiments, we \textit{only} induce sparsity in the feed-forward layers of both encoder and decoder blocks (\autoref{table:compute_contribution} shows that feed-forward layers account for around 64\% of FLOPS and 67\% of parameters of the entire model. Therefore, we focus on feed-forward layers for sparsification). 
We follow the standard practice of fine-tuning
using the final learning rate used during the original training phase~\cite{liu2018rethinking}.

\niparagraph{Training details.}
Table~\ref{table:setup} shows the details of training hyperparameters that we use for all the evaluations.
For each experiment, we use a TPUv3 with 32 cores.
\begin{table}[!htp]\centering
\caption{Model configurations and hyper-parameters.}

\begin{tabular}{lrr}\toprule
number of encoder layers &6 \\
number of decoder layer &6 \\
hidden dimension size &1024 \\
feed forward dimension size &4096 \\
number of head &16 \\
max sequence length &256 \\
training set &WMT-17 \\
testing set &WMT-14 \\
learning rate &0.0625 \\
warmup steps &1000 \\
decay factor &0.5 \\
batch size &512 \\
training steps &200K \\
Adam optimizer &beta1 = 0.9, beta2 = 0.92 \\
\bottomrule
\end{tabular}
\label{table:setup}
\end{table}
\subsection{Sparsification Method Baseline}
\label{sec:method}
\niparagraph{Fine-grained N:M sparsity.}
We follow the proposed method in~\cite{sr_ste} to induce structured N:M sparsity from scratch, as shown in \autoref{fig:alg_flow}(a). 
This method employs standard online magnitude-based pruning with an introduced sparse-refined regularization term.  
This regularization term applies refined gradients for pruned weights during backward pass.
The authors use the refined gradient updates to increase the likelihood of pruning the same network weights at each training step, which purportedly leads to a more robust sparse training.
While this work proposes to re-evaluate the pruning mask after each training iteration, we find this process time-consuming which significantly slows down the training process on TPU.
Therefore, we moderately alter the frequency of updating the pruning mask to 1000 training steps. 
We ablate the importance of the frequency of updating pruning masks.
Our results show that the model accuracy for WMT task is not sensitive to this parameter, as shown in \autoref{table:ab_update_freq}.

\begin{table}[ht]\centering
\caption{The effect of update frequency in SR-STE~\cite{sr_ste}. Raising the ``Update Frequency'' increases the training time significantly. Hence, we only include the results for ``Update Frequency'' 100 and 1000.}
\scriptsize
\begin{tabular}{lrrrr}\toprule
\multicolumn{2}{c}{\multirow{2}{*}{Accuracy}} &\multicolumn{2}{c}{Update Frequency} \\\cmidrule{3-4}
& &every 1000 steps &every 100 steps \\\midrule
\multirow{4}{*}{Sparsity Target} &1:16 &0.709 &0.710 \\
&1:32 &0.707 &0.707 \\
&1:64 &0.706 &0.706 \\
&1:128 &0.706 &0.706 \\
\bottomrule
\end{tabular}
\label{table:ab_update_freq}
\end{table}

\subsection{Proposed Sparsification Methods}
In this section, we propose two sparsification methods that employ a decaying mechanism to gradually induce the target sparsity on the model.
Note that, we do not alter the gradient update rule in either of these proposed methods.
Instead, we simply employ various gradual update rules to the pruning mask itself.

\niparagraph{Pruning mask decay.}
In the first approach, instead of using a binary pruning mask (e.g. ``0'' indicates pruning locations), we use a floating-point pruning mask with decaying, as shown in \autoref{fig:alg_flow}(b).
At the start of training, we employ an all-ones matrix as the pruning mask that simply indicates no pruning.
At the beginning of sparse training phase, we use the same standard online magnitude-based pruning criteria to identify the locations of pruned weights.
However, in contrast to prior work in which ``0'' is used to prune the weights, we use $0 < \beta < 1.0$ for the pruned weights in the pruning mask.
We gradually decrease the value of $\beta$ at different intervals following the formula $\beta^{d}$, where $d$ indicates the decaying iteration index (e.g. $\beta^1$, $\beta^2$, $\beta^3$, ...).
After sufficient decaying intervals, we set $\beta$ to zero to indicate the locations of pruned weights.
We postulate that using a non-binary pruning mask enables the gradients of pruned weights to flow through the network leading to a more robust sparse training and better model performance.

\niparagraph{Sparse structure decay.}
In the second proposed sparsification method, we apply a decaying mechanism on the structure of pruning mask, gradually increasing sparsification degree.
At the beginning of sparse training phase, we start with M-1:M structured sparsity.
As training progresses, we increase the sparsification degree by applying $\frac{M}{2^d}:M$ structured sparsity at different decaying intervals. Similar to previous method, $d$ denotes the decaying iteration index.
This method at its crux follows a similar hypothesis as the pruning mask decay. That is, enabling the gradient of pruned weights to flow through the network. However, because we still use a binary pruning mask, the contribution of the gradients of the pruned weights to the network reduces after each decaying interval.

\section{Evaluation}
\label{sec:eval}

\subsection{Methodology}

\niparagraph{Task.} We use translation as our target task. We use WMT dataset (En-De)~\cite{wmt_dataset}. 

\niparagraph{Comparisons.}
In this study, we compare the effectiveness of different SOTA sparsification methods.
All methods train for $n$ steps.

\squishlist
\item \textbf{Dense:} Dense training without sparsification for $n$ steps (\autoref{fig:dense_sparse}(e)).
\item \textbf{Dense-sparse:} Dense train for $d$ steps, sparsify, and fine-tune for ($n$-$d$) steps, as in ~\cite{mishra2021accelerating} (\autoref{fig:dense_sparse}(f)). 
\item \textbf{Sparse:} SR-STE~\cite{sr_ste}-based sparse training for $n$ steps (\autoref{fig:alg_flow}(a)).

\item \textbf{Structure Decay:} Dense train for $d$ steps, structure decay the sparsity pattern for ($n$-$d$-$s$) steps, and fine-tune for $s$ steps (\autoref{fig:dense_sparse}(g)). The structure decay is set to decay by the power of 2 (\autoref{sec:method}). For example, when target sparsity pattern is 1:16, we divide ($n$-$d$-$s$) steps to five equal time frame, and the sparsity pattern of each time frame is 15:16, 8:16, 4:16, 2:16, and 1:16, respectively.  
\item \textbf{Mask Decay:} Dense train for $d$ steps, mask decay the sparsity pattern for ($n$-$d$-$s$) steps, and fine-tune for $s$ steps (\autoref{fig:alg_flow}(b)). We use the mask decay rate ($\beta$) of 0.9 and mask update period of 1000 steps. 
In the above experiments, we use $n$ = 200K, $d$ = 20K, $s$ = 20K.
\squishend

\subsection{Comparing with Baseline}

\niparagraph{Quality.}
We compare Structure Decay and Mask Decay, with two baseline Dense-Sparse~\cite{mishra2021accelerating} and Sparse~\cite{sr_ste} in \autoref{table:overall}. We evaluate the methods on different sparsity targets. Sparsity target is the final sparsity pattern we will achieve after model training. For example, sparsity target of 1:32 means the trained model will have only 1 non-zero parameters every 32 parameters. \autoref{table:overall} shows that Dense-Sparse performs similarly to Sparse, and Mask Decay achieves the best accuracy across all sparsity targets. Structure Decay performs the second best. More interestingly,
Mask Decay can help achieve similar or better accuracy than the ``unstructured sparsity'' ones.

\emph{Our results indicate that the ``Mask Decay'' pruning method on dense layers enables models to be pruned structurally while achieving comparable or even better accuracy to ``unstructured pruning''.}

\begin{table}[!htp]\centering
\caption{Comparisons between different sparsification strategies.}
\scriptsize
\begin{tabular}{lr|r|rrrrrrr}\toprule
\multicolumn{2}{c|}{Accuracy} &Dense &\multicolumn{4}{c}{Structure Sparsity} &\multicolumn{2}{c}{Unstructure Sparsity} \\\cmidrule{1-9}

\multicolumn{2}{c|}{Schedule} &Dense &Structure Decay &Mask Decay &Dense-Sparse~\cite{mishra2021accelerating} &Sparse~\cite{sr_ste} &Dense-Sparse &Unstr Sparse \\\midrule
\multirow{4}{*}{\makecell{Sparsity\\ Target}} &1:16 &0.747 &\textbf{0.717} &\textbf{0.717} &0.714 &0.709 &0.714 &0.714 \\
&1:32 &0.747 &0.713 &\textbf{0.714} &0.710 &0.707 &0.711 &0.712 \\
&1:64 &0.747 &0.710 &\textbf{0.711} &0.708 &0.707 &0.711 &0.711 \\
&1:128 &0.747 &0.708 &\textbf{0.711} &0.708 &0.707 &0.708 &0.709 \\

\bottomrule
\end{tabular}
\label{table:overall}
\end{table}

\niparagraph{Performance.}
Note that there are no off-the-shelf accelerators that can support 1:16 or more aggressive sparsity patterns. 
To demonstrate the potential performance, we build a cost model to estimate the FLOPS and memory sizes. \autoref{table:cpr_flops} shows that after 1:16 sparsification, the model sizes will reduce by 62\% and inference FLOPS will reduce by 60\%. More interestingly, since different methods have different sparsification schedules, the averaged training FLOPS across their training time will be different. ``Dense-Sparse'' is the most straight-forward and light-weighted training schedule in terms of FLOPS. ``Sparse'' relies on continuous mask update across the full training steps (\autoref{fig:alg_flow}(a)), therefore performing worse in terms of FLOPS. Structure Decay becomes the most performant method to achieve the best quality at 1:16 sparsity. However, to sparsify more aggressively, Mask Decay is still the best method (\autoref{table:overall}).

\begin{table}[!htp]\centering
\caption{Comparisons of model performance (FLOPS) and quality (accuracy).}
\scriptsize
\begin{tabular}{lrrrrrr}\toprule
Sparsity Target (1:16) &Dense &Structure Decay &Mask Decay &Dense-Sparse &Sparse \\\midrule
Params (MB) &151.0 &56.6 &56.6 &56.6 &56.6 \\
Inference TFLOPS &41.2 &16.5 &16.5 &16.5 &16.5 \\
Training TFLOPS &123.7 &108.0 &121.2 &\textbf{101.4} &123.7 \\
Accuracy &0.75 &\textbf{0.72} &\textbf{0.72} &0.714 &0.709 \\
\bottomrule
\end{tabular}
\label{table:cpr_flops}
\end{table}

\subsection{Ablation Studies}
\label{sec:ablations}
\niparagraph{Dense training v.s. training from scratch for SR-STE.}
\label{sec:ab_srste}
SR-STE uses sparse training from scratch. All the other methods that we evaluated have a dense training phase at the first few steps (epochs). This recipe has been proven to be effective as shown in the previous experiments and many prior works~\cite{mishra2021accelerating,pool2021channel, frankle2018lottery,lee2018snip,evci2019difficulty,han2015deep, guo2016dynamic,he2017channel,molchanov2016pruning,yao2019balanced,zhu2017prune,gamboa2020campfire,narang2017exploring,narang2017block,elsen2020fast,park2018squantizer,kalchbrenner2018efficient,evci2020rigging}. Therefore, we experiment on adding a dense training phase at the beginning of SR-STE training, as shown in \autoref{table:ab_srste}. We found that adding few steps of dense training (1.25\% - 10\% of the total training steps) can increase the accuracy by around 0.002 to 0.003. This tells that few steps of dense training does help achieve better performance even for SR-STE. Interestingly, the improved SR-STE becomes competitive to the proposed Structure Decay. However, Mask Decay is still consistently better.

\begin{table}[!htp]\centering
\caption{Ablation: SR-STE augmented with few epochs of dense training.}
\scriptsize
\begin{tabular}{lrrrrrrr}\toprule
&\multicolumn{2}{c}{Training Schedule} &\multicolumn{4}{c}{Accuracy} \\\cmidrule{2-7}
&Dense steps &SR-STE -styled Sparse steps &1:16 &1:32 &1:64 &1:128 \\\cmidrule{2-7}
SR-STE &0 &200K &0.710 &0.707 &0.706 &0.706 \\\midrule
\multirow{4}{*}{Dense + SR-STE} &2.5K &197.5K &0.712 &0.710 &\textbf{0.708} &0.706 \\
&5K &195K &0.712 &0.709 &0.707 &\textbf{0.708} \\
&10K &190K &0.712 &\textbf{0.710} &0.707 &\textbf{0.708} \\
&20K &180K &\textbf{0.713} &\textbf{0.710} &\textbf{0.708} &0.707 \\
\bottomrule
\end{tabular}
\label{table:ab_srste}
\end{table}

\niparagraph{Effect of dense training steps ($d$).}
\label{sec:ab_d_steps}
Both our proposed methods include a dense training phase. We do an ablation study on different number of dense training steps in \autoref{table:ab_d_steps}. We found that changing the dense step between 1.25\% - 10\% of the total training steps does not observably change the accuracy performance. However, empirically, we found that dense training phase is still essential. The model cannot achieve as competitive accuracy without few epochs of dense training. 

\begin{table}[!htp]\centering
\caption{Ablation: The effect of number of dense training steps ($d$).}
\scriptsize
\begin{tabular}{lrrrrrrrrrr}\toprule
\multicolumn{2}{c}{Accuracy} &\multicolumn{4}{c}{Mask Decay} &\multicolumn{4}{c}{Structure Sparsity} \\\cmidrule{1-10}
\multicolumn{2}{c}{Sparsity Target} &1:16 &1:32 &1:64 &1:128 &1:16 &1:32 &1:64 &1:128 \\\midrule
\multirow{4}{*}{Dense steps (d)} &2.5 K &0.7155 &0.7134 &0.7106 &0.7100 &0.7157 &0.7134 &0.7108 &0.7106 \\
&5 K &\textbf{0.7160} &0.7127 &\textbf{0.7110} &0.7093 &0.7160 &0.7136 &\textbf{0.7117} &0.7100 \\
&10 K &0.7157 &\textbf{0.7137} &0.7103 &0.7094 &0.7164 &\textbf{0.7141} &0.7107 &0.7098 \\
&20 K &0.7156 &0.7126 &0.7107 &\textbf{0.7104} &\textbf{0.7165} &0.7128 &0.7115 &\textbf{0.7107} \\
\bottomrule
\end{tabular}
\label{table:ab_d_steps}
\end{table}

\niparagraph{Effects of fine-tuning steps ($s$).}
\label{sec:ab_s_steps}
We also have a sets of study on number of fine-tuning steps in \autoref{table:ab_s_steps}. We found that for both of our propose methods the fine-tuning steps between 10\% - 20\% of the total training steps does not observably change the accuracy performance. However, empirically, we also found few steps of fine-tuning at the end is essential to recover the accuracy. 

\begin{table}[!htp]\centering
\caption{Ablation: The effect of number of fine-tuning steps ($s$).}
\scriptsize
\begin{tabular}{lrrrrrrrrrr}\toprule
\multicolumn{2}{c}{Accuracy} &\multicolumn{4}{c}{Mask Decay} &\multicolumn{4}{c}{Structure Sparsity} \\\cmidrule{1-10}
\multicolumn{2}{c}{Sparsity Target} &1:16 &1:32 &1:64 &1:128 &1:16 &1:32 &1:64 &1:128 \\\midrule
\multirow{2}{*}{Fine-tuning steps (s)} &20 K &0.7153 &0.7130 &\textbf{0.7107} &\textbf{0.7098} &\textbf{0.7160} &\textbf{0.7125} &\textbf{0.7095} &\textbf{0.7072} \\
&40 K &\textbf{0.7161} &\textbf{0.7132} &0.7106 &0.7097 &0.7121 &0.7093 &0.7081 &0.7065 \\
\bottomrule
\end{tabular}
\label{table:ab_s_steps}
\end{table}

\niparagraph{Effects of $\beta$ in mask decay.}
\label{sec:ab_decay_rate}
Note that the extreme case of a mask decay rate ($\beta$=0) will turn the pruning mask back to the conventional 1/0 mask. As shown in \autoref{table:ab_decay_rate}, we found a mask decay rate of 0.9 is better than an aggressive one (0.001). It tells the Mask Decay technique does contribute and lead to better accuracy performance. 
\begin{table}[!htp]\centering
\caption{Ablation: The effect of mask decay rate ($\beta$).}
\scriptsize
\begin{tabular}{lrrrrrr}\toprule
\multicolumn{2}{c}{Accuracy} &\multicolumn{4}{c}{Mask Decay} \\\cmidrule{1-6}
\multicolumn{2}{c}{Sparsity Target} &1:16 &1:32 &1:64 &1:128 \\\midrule
\multirow{2}{*}{Mask decay rate ($\beta$)} &0.9 &\textbf{0.715} &\textbf{0.713} &\textbf{0.711} &\textbf{0.710} \\
&0.001 &0.712 &0.709 &0.708 &0.707 \\
\bottomrule
\end{tabular}
\label{table:ab_decay_rate}
\end{table}
\section{Limitations}

This paper studies only translation task with one sparsification recipe (Criteria: magnitude pruning, Schedule: Structure Decay or Mask Decay, Pattern N:M structured pruning) and only prunes feed-forward layers. 1) We show we can effectively prune the most compute- and parameter-heavy layers in our model, feed forward layers. An interesting next-step is to prune other layers such as projections layers as well to further compress the model. 2) Studies on other language tasks or visions tasks and on different models such as on Resnet~\cite{resnet} or ViTs~\cite{vit}) would be an interesting follow-up. 3) In addition, we only study one combination of sparsification recipe. We might discover better recipe by exploring other combinations such as salient-based pruning + Mask Decay + N:M structured pruning, magnitude pruning + Structure Decay + unstructured sparsity, or many others. 4) Lastly, in the evaluations, most of the hyper-parameters are set manually. We did a limited scope of hyper-parameters sweep in \autoref{sec:ablations}. A full-fledged hyper-parameter search might discover more performance improvement in Mask Decay and Structure Decay.

\section{Conclusion}
In this work, we study and evaluate various training recipes for N:M structured sparsity.
Building on this study, we propose and compare two new training recipes for N:M structured sparsity based on decaying mechanisms.
We study the trade-off between model accuracy and training compute cost (FLOPs) across these training recipes.
We show that gradual decay of pruning mask values consistently yield better model accuracy, on-par with unstructured sparsity, on translate task at the cost of modest increase in the training compute cost.
While structured sparsity seems to be better positioned for hardware acceleration, its associated training cost should not be overlooked.
This work represents a first step in evaluating training recipes for structured sparsity from the perspective of trade-off between model accuracy and compute cost.
As future work, we plan to expand the pool of models to other tasks and models and develop a platform to systematically evaluate and compare various sparse training recipes both for model accuracy and training cost.

\section*{Acknowledgements}
We would like to extend our gratitude towards Jeremiah Willcock, Penporn Koanantakool, Chandu Thekkath, and our extended team at Google Research, Brain Team.
%

\bibliographystyle{plain}
\bibliography{reference}

\end{document}